\documentclass[conference]{IEEEtran}
\IEEEoverridecommandlockouts
\usepackage{cite}
\usepackage{amsmath,amssymb,amsfonts}
\usepackage{algorithmic}
\usepackage{graphicx}
\usepackage{textcomp}
\usepackage{xcolor}
\usepackage{multirow}
\usepackage{makecell}
\usepackage{colortbl}
\usepackage{booktabs}
\usepackage{xspace}
\definecolor{cvprblue}{rgb}{0.21,0.49,0.74}

\usepackage[breaklinks,colorlinks,allcolors=cvprblue]{hyperref}
\usepackage[capitalize]{cleveref}
\crefname{section}{Sec.}{Secs.}
\Crefname{section}{Section}{Sections}
\Crefname{table}{Table}{Tables}
\crefname{table}{Tab.}{Tabs.}

\def\BibTeX{{\rm B\kern-.05em{\sc i\kern-.025em b}\kern-.08em
    T\kern-.1667em\lower.7ex\hbox{E}\kern-.125emX}}

\newcommand{\ie}{\textit{i.e.}\@\xspace}

\begin{document}

\title{Efficient Explicit Joint-level Interaction Modeling with Mamba for Text-guided HOI Generation}

\author{
Guohong Huang$^\dagger$, Ling-An Zeng$^{\dagger}$,  
Zexin Zheng, Shengbo Gu, Wei-Shi Zheng*\\
Sun Yat-sen University, China \\
Key Laboratory of Machine Intelligence and Advanced Computing, Ministry of Education, China \\
{\tt\small \{huanggh37, zenglan3, zhengzx25, gushb3\}@mail2.sysu.edu.cn, wszheng@ieee.org}

}

\maketitle
\def\thefootnote{}\footnotetext{$\dagger$ Equal contribution; * Corresponding authors.}

\vspace{-0.5cm}
\begin{abstract}
We propose a novel approach for generating text-guided human-object interactions (HOIs) that achieves explicit joint-level interaction modeling in a computationally efficient manner. Previous methods represent the entire human body as a single token, making it difficult to capture fine-grained joint-level interactions and resulting in unrealistic HOIs. However, treating each individual joint as a token would yield over twenty times more tokens, increasing computational overhead.
To address these challenges, we introduce an Efficient Explicit Joint-level Interaction Model (EJIM). EJIM features a Dual-branch HOI Mamba that  separately and efficiently models spatiotemporal HOI information, as well as a Dual-branch Condition Injector for integrating text semantics and object geometry into human and object motions. Furthermore, we design a Dynamic Interaction Block and a progressive masking mechanism to iteratively filter out irrelevant joints, ensuring accurate and nuanced interaction modeling.
Extensive quantitative and qualitative evaluations on public datasets demonstrate that EJIM surpasses previous works by a large margin  while using only 5\% of the inference time. Code is available \href{https://github.com/Huanggh531/EJIM}{here}.
% The code will be public.
\end{abstract}


\section{Introduction}
\label{sec:intro}
Text-guided human-object interaction (HOI) generation aims to produce smooth and realistic motions of both humans and objects conditioned on text descriptions, which is critical for automatic character animation, robotics, and embodied AI. While recent works \cite{hoianimator, cghoi, chois, interdiff,omomo,hoidiff} have made progress, these methods fail to achieve explicit joint-level interaction modeling. This limitation arises because they treat the entire human body as a single token and model interactions solely between the human and object tokens. \textit{We argue that explicit joint-level interaction modeling is essential for HOI, as it allows the model to capture detailed relationships between human joints and objects, such as relative positions and contact distances.}

To achieve joint-level interaction modeling, an intuitive approach is to input all joints (both human joints and objects) into a Transformer \cite{transformer} for interaction modeling. However, considering the motion length and the number of human joints (23 joints), the computational complexity of this method is approximately 400x greater than non-joint-level interaction modeling due to the $O(n^2)$ complexity of self/cross-attention mechanisms. An alternative is the recently emerged Mamba \cite{mamba}, which offers comparable modeling capabilities to Transformers while using linear scalability with sequence length.

\begin{figure}[t]
    \centering
    \includegraphics[width=\linewidth]{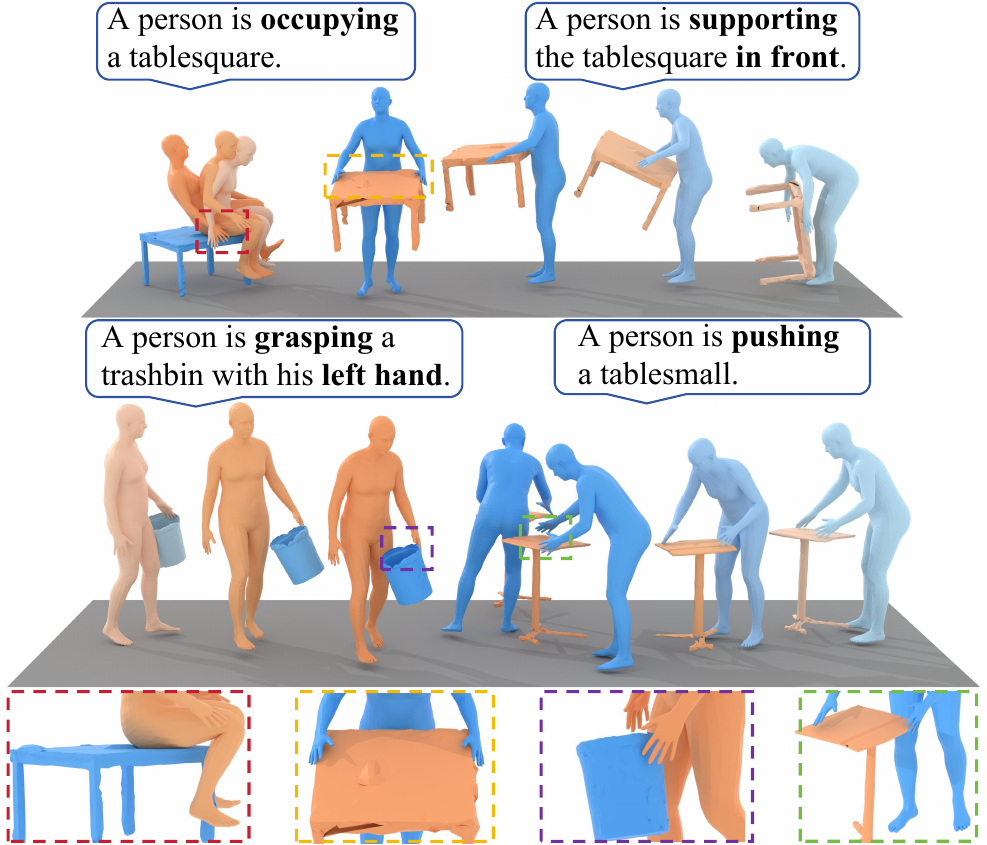}
    \vspace{-0.65cm}
    \caption{Our EJIM can generate realistic 3D human-object interactions guided by text descriptions and object geometry, with colors transitioning from lighter to darker to represent the passage of time.}
    \vspace{-0.6cm}
    \label{fig:enter-label}
\end{figure}

Based on the above motivations, we propose a novel \textbf{\underline{E}fficient Explicit \underline{J}oint-level \underline{I}nteraction \underline{M}odel} (EJIM) that achieves explicit joint-level interaction modeling in an efficient manner. Our EJIM first introduces two novel blocks: a \textbf{Dual-branch HOI Mamba} (DHM) and a \textbf{Dual-branch Condition Injector} (DCI). The DHM separately models spatiotemporal HOI information in a dual-branch manner. Specifically, DHM adopts a novel limb-guided spatial scan to enhance relation capturing between human joints guided by limb partitioning. Moreover, our DCI introduces a specialized scan to inject text semantics and object geometry into the human and object motions in a dual-branch manner. In this way, our EJIM can efficiently model spatiotemporal HOI information and inject conditions into HOI sequences at the joint level.

Furthermore, to enhance interaction modeling between human joints and the object, our EJIM introduces a novel \textbf{Dynamic Interaction Block} (DIB) along with a progressive masking mechanism. Our DIB accurately models the interaction between the object and specific human joints based on the attention mechanism. Moreover, considering that most human joints do not interact with the object, DIB adopts a progressive masking mechanism that progressively filters out irrelevant human joints to achieve accurate interaction modeling. 
% In this way, our DIB successfully achieves explicit accurate joint-object interaction modeling.

We quantitatively and qualitatively demonstrate the effectiveness and efficiency of our EJIM on the BEHAVE\cite{behave} and OMOMO \cite{omomo} datasets. Results show that EJIM surpasses existing methods by a large margin in \textbf{motion generation quality} and  \textbf{interaction quality} while using only \textbf{5\% of the inference time}. We also provide extensive ablation studies and visualizations to aid understanding.

\section{Related Works}
\noindent\textbf{Human-object Interaction Generation.}
Some methods \cite{hassan2023synthesizing, merel2020catch} aim to simulate physical rules using reinforcement learning but are often limited in generalization and challenging to train. Other approaches employ generative models to produce HOIs. Several works \cite{Synthesizing, yi2024generating} focus on generating interactions between humans and static objects, such as in various scenes. In contrast, some studies \cite{text2hoi, cams} explore interactions between hands and dynamic objects. Furthermore, certain methods \cite{omomo, interdiff, hoidiff,chainhoi, chois} concentrate on generating interactions between the full body and dynamic objects. For example, InterDiff \cite{interdiff} introduces interaction diffusion and correction mechanisms to generate and refine HOIs. HOI-Diff \cite{hoidiff} incorporates an affordance predictor to estimate contact areas and refine contact distances. Beyond text descriptions, CHOIS \cite{chois} utilizes object waypoints to control the generation process. In this work, we focus on generating interactions between the full body and dynamic objects guided by texts. Unlike existing methods, our method achieves efficient explicit joint-level interaction modeling and introduces a progressive masking mechanism for more accurate interaction modeling.

% \vspace{0.1cm}
\noindent\textbf{State Space Models.}
State Space Models (SSMs)  \cite{kalman1960new} are a class of control systems for processing continuous inputs. The Structured State Space Sequence model \cite{ssm1, ssm2} proposes a discretization strategy to efficiently process discrete inputs, which captures complex dependencies while maintaining linear scalability with sequence length. The recent Mamba \cite{mamba}, an advanced variant, demonstrates modeling capabilities comparable to Transformers \cite{transformer}. Due to its potential in sequence modeling, Mamba has revolutionized various domains, including image recognition \cite{xu2024survey} and motion generation \cite{zhang2025motion}. In this work, we leverage Mamba to realize efficient and effective text-guided HOI generation.

\section{Approach}
\vspace{-0.2cm}
\subsection{Preliminary}
\noindent\textbf{Mamba.}
Mamba \cite{mamba} is built upon State Space Models (SSMs) \cite{kalman1960new}, which transform an input $x(t) \in \mathbb{R}$ to an output $y(t) \in \mathbb{R}$ via a hidden state $h(t) \in \mathbb{R}^N$, governed by parameters $\mathbf{A} \in \mathbb{R}^{N \times N}$, $\mathbf{B} \in \mathbb{R}^{N \times 1}$, $\mathbf{C} \in \mathbb{R}^{1 \times N}$, and $D \in \mathbb{R}$:
\begin{equation}
    h'(t) = \mathbf{A} h(t) + {\mathbf{B}} x(t), \; \; \;
    y(t) = \mathbf{C} h(t) + D x(t).
\end{equation}
A zero-order hold is proposed to process discrete inputs. The discrete SSM can be  efficiently computed by a convolution process. After that, Mamba propose a selective mechanism by using input-dependent parameters $\mathbf{B}=\mathbf{S}_B(x)$, $\mathbf{C}=\mathbf{S}_C(x)$, and $\Delta = \mathbf{S}_\Delta (x)$, to build a time-varying model for modeling 
complex inputs while maintaining computational tractability.

% \vspace{0.1cm}
\noindent\textbf{Diffusion Models.}
Diffusion models \cite{ho2020denoising} are generative frameworks that operate by gradually adding Gaussian noise to a sample through a series of timesteps, and then training a model to reverse this process, ultimately recovering the original signal. 
Formally, given a data distribution $q(\mathbf{x}_0)$, the forward diffusion process is defined as a Markov chain $\mathbf{x}_0 \rightarrow \mathbf{x}_1 \rightarrow \cdots \rightarrow \mathbf{x}_T$, where each step adds Gaussian noise:
\begin{equation}
    q(\mathbf{x}_t | \mathbf{x}_{t-1}) = \mathcal{N}(\mathbf{x}_t; \sqrt{1-\beta_t \mathbf{x}_{t-1}, \beta_t I}),
\end{equation}
with $\beta_t$ controlling the noise schedule. The model then learns the reverse process to iteratively remove noise.

\subsection{Problem Formulation}
Text-guided HOI generation aims to produce a 3D HOI sequence $\mathbf{x} = \{\mathbf{x}^h, \mathbf{x}^o\}$ of length $L$, conditioned on a given text description $S$ and object geometry $G$. To explicitly model interactions at the joint level, we introduce a new HOI representation scheme that preserves information for each joint.

Specifically, each non-root human joint is represented as a 12-dimensional vector that includes its local joint positions, velocities, and rotations in the root space, inspired by the redundant motion representations used in text-driven motion generation \cite{t2m}. The root joint is represented as a 7-dimensional vector encoding its global position, angular velocity about the Y-axis, linear velocities on the XZ-plane, and height. For the object, a 6-dimensional vector is used to capture its global rotation and translation. Moreover, to mitigate foot skating, we add a virtual joint represented by a 4-dimensional binary vector indicating foot–ground contact. 

For consistency, all joints and object representation vectors are zero-padded to a dimensionality of 12. Thus, the 3D HOI sequence is denoted as $\mathbf{x} = \{\mathbf{x}^h \in \mathbb{R}^{L \times J \times \hat{D}}, \mathbf{x}^o \in \mathbb{R}^{L \times \hat{D}}\}$, where $J$ is the number of joints, including the virtual joint.

\begin{figure*}
    \centering
    \includegraphics[width=\linewidth]{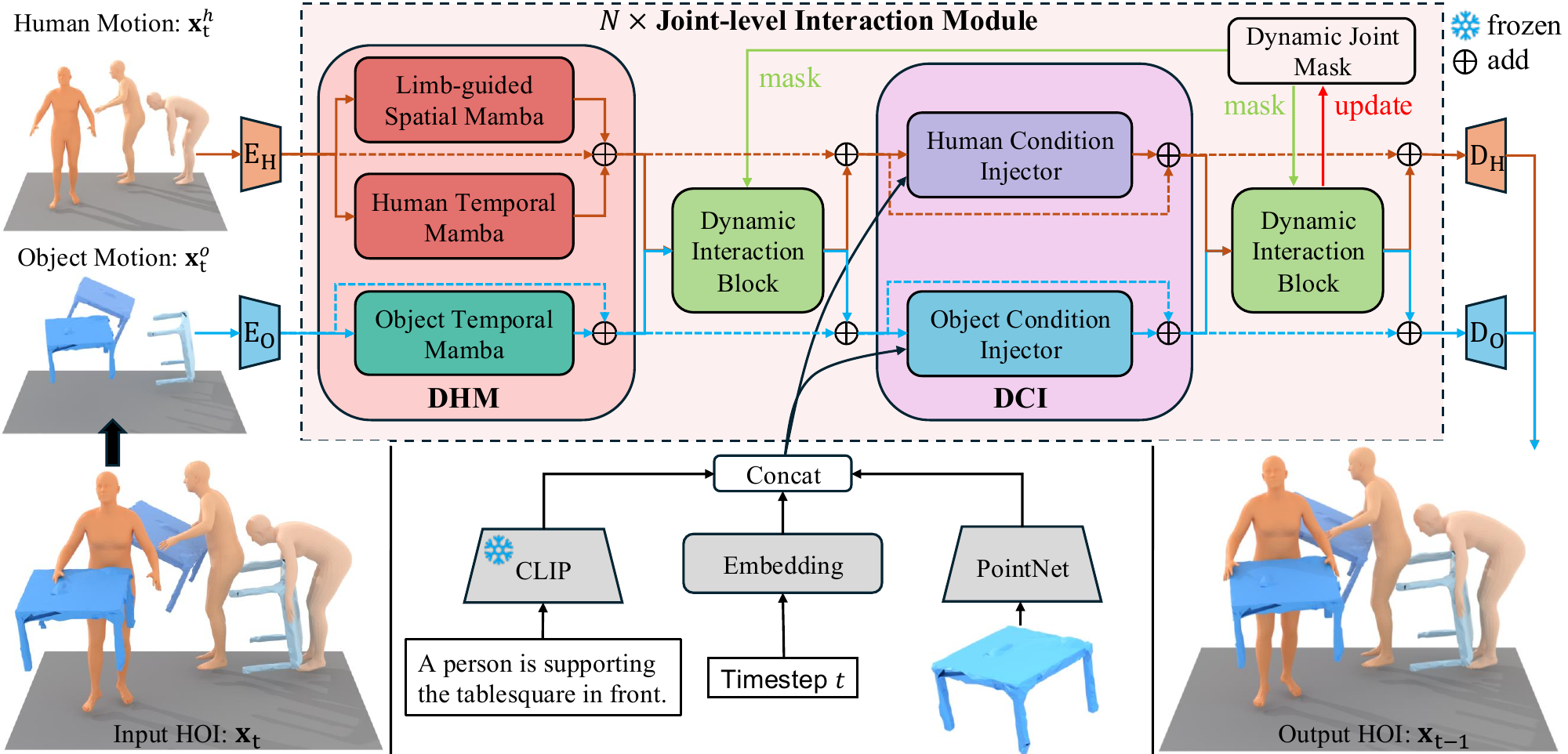}
    \vspace{-0.7cm}
    \caption{Overview of our EJIM.  Our EJIM takes a noisy HOI sequence $\mathbf{x}_t = \{\mathbf{x}_t^o, \mathbf{x}_t^h\}$ as input and generates denoised results $\mathbf{x}_{t-1}$.  $\mathbf{x}_t^o$ and $\mathbf{x}_t^h$ are projected into a latent space via linear projections $\text{E}_o$ and $\text{E}_h$, respectively. In each module, a Dual-branch HOI Mamba (DHM) is used to model spatiotemporal information, while a Dual-branch Condition Injector (DCI) injects conditional information. Two Dynamic Interaction Blocks are employed to model interactions, guided by a dynamic interaction mask that is progressively updated in each module to filter out irrelevant joints, enabling more accurate interaction modeling.}
    \vspace{-0.4cm}
    \label{fig:overview}
\end{figure*}

\subsection{Overview of EJIM}
% Unlike existing methods \cite{hoidiff, hoianimator, cghoi, chois, interdiff,omomo} that are unable to explicitly model relationships between human joints and objects,
Unlike existing methods \cite{hoidiff, hoianimator, cghoi, chois, interdiff,omomo}, we propose a novel \textbf{\underline{E}fficient Explicit \underline{J}oint-level \underline{I}nteraction \underline{M}odel} (EJIM) that enables efficient and explicit joint-level interaction modeling. As illustrated in \cref{fig:overview}, EJIM is a diffusion-based method that takes a noisy HOI sequence $\mathbf{x}_t$ as input and predicts its denoised counterpart $\mathbf{x}_{t-1}$. The human motion $\mathbf{x}_t^h$ and object motion $\mathbf{x}_t^o$ are first encoded into latent space using separate encoders ($\text{E}_\text{H}$ and $\text{E}_\text{O}$). The final outputs are decoded back into the original space by corresponding decoders ($\text{D}_\text{H}$ and $\text{D}_\text{O}$). All these encoders and decoders ($\text{E}_\text{H}$, $\text{E}_\text{O}$, $\text{D}_\text{H}$, and $\text{D}_\text{O}$) are implemented as linear layers. Additionally, we leverage a PointNet \cite{pointnet} to extract object geometry features and a fixed pretrained CLIP model \cite{clip} to derive text features.

In detail, EJIM consists of $N$ identical Joint-level Interaction Modules, each containing a Dual-branch HOI Mamba (DHM), a Dual-branch Condition Injector (DCI), and two Dynamic Interaction Blocks (DIB). The DHM first models spatiotemporal information for both human and object motions in a dual-branch manner. Its outputs are then fed into the first DIB, which explicitly models interactions between potential interaction joints and the object, using the joint mask to exclude irrelevant joints. The DCI injects conditioning information into the human and object feature branches. Subsequently, the second DIB further refines interaction modeling and updates the joint mask, progressively filtering out non-contributing joints.

\subsection{Dual-branch HOI Mamba}
To model spatiotemporal information for both human and object motion, we propose a Dual-branch HOI Mamba (DHM), consisting of a human branch and an object branch. Both branches are implemented in Mamba blocks \cite{mamba} for efficiently  capturing complex dependencies.

To capture inter-frame temporal information, the input human motion features $\mathbf{h} \in \mathbb{R}^{L \times J \times D}$ and object features $\mathbf{o} \in \mathbb{R}^{L \times D}$ are fed into a Human Temporal Mamba and an Object Temporal Mamba, respectively. We independently model inter-frame temporal information for each joint’s features $\mathbf{h}^j \in \mathbb{R}^{L \times D}$. Note that $D$ represents the model dimension.

\begin{figure}[ht]
    \centering
    \vspace{-0.1cm}
    \includegraphics[width=\linewidth]{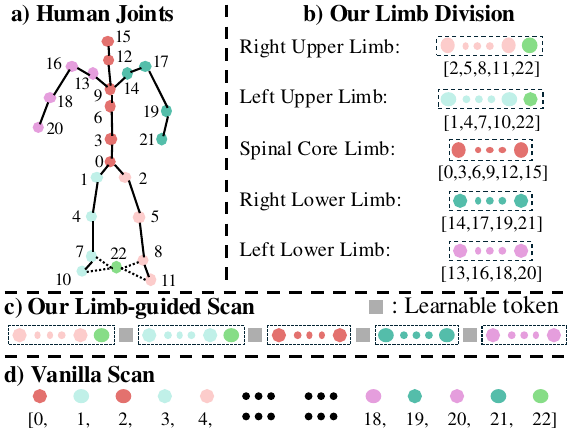}
    \vspace{-0.7cm}
    \caption{ (a) Illustration of human joints. (b) Our limb division scheme. Here, the virtual foot-ground contact joint is duplicated and assigned to both lower limbs to mitigate foot skating. (c) The Limb-guided scan in our Spatial Mamba reorders joints by limb groupings and inserts learnable tokens to define distinct limbs. (d) The vanilla scan approach for comparison.}
    \vspace{-0.65cm}
    \label{fig:lgm}
\end{figure}

To capture intra-frame spatial information for human motion features, we propose a novel Limb-guided Spatial Mamba. As shown in subfigure (c) of \cref{fig:lgm}, given the joint features $\mathbf{h}_l \in \mathbb{R}^{J \times D}$ at the $l$-th frame, we reorder the joints according to their limb groupings and insert learnable tokens to delineate these distinct limb joint groups. Unlike the vanilla approach, which simply feeds all joints into Mamba without preprocessing, the Limb-guided Spatial Mamba enables the model to become aware of limb divisions, thereby facilitating more effective relational modeling, as demonstrated in \cref{sec:ablation}.

\vspace{-0.1cm}
\subsection{Dual-branch Condition Injector}
While our DHM captures the motion of human joints and the object, it does not incorporate external conditions. To address this limitation, we propose a Dual-branch Condition Injector (DCI), which injects various conditions—including text descriptions, diffusion timesteps, and object geometry—into both human and object motion streams, thereby enabling controllable HOI generation. Similar to DHM, our DCI consists of two branches, namely Human Condition Injector and Object Condition Injector, each dedicated to integrating conditions into the respective motion sequences.

In detail, both branches are implemented in Mamba blocks. Since Mamba does not include cross-attention or similar operations to gather information from additional sequences, we concatenate condition tokens $\mathbf{c} \in \mathbb{R}^{N \times D}$ with the human or object tokens before feeding them into Mamba blocks: 
\begin{equation} 
    \hat{\mathbf{m}} = \text{Mamba}(\text{Concat}([\mathbf{c}, \mathbf{m}])), 
\end{equation} 
where $\mathbf{m}$ represents either human joint motion $\mathbf{h}^j \in \mathbb{R}^{L \times D}$ or object motion $\mathbf{o} \in \mathbb{R}^{L \times D}$. Within Mamba blocks, $\mathbf{m}$ assimilates information from $\mathbf{c}$. After processing, we retain only motion-related tokens in $\hat{\mathbf{m}}$ and discard condition tokens.

% \vspace{-0.1cm}
\subsection{Dynamic Interaction Block}
Furthermore, we introduce a Dynamic Interaction Block (DIB) with a dynamic joint mask for precise interaction modeling. Conditioned on this mask, DIB captures interactions between selected human joints and the object. After each Joint-level Interaction Module, the mask is progressively updated to filter out irrelevant joints. To improve this filtering process, DIB leverages relational modeling among all visible joints, ensuring more accurate removal of irrelevant joints.

The DIB employs an attention mechanism to model interactions. At the $l$-th frame, we first concatenate the object token $\mathbf{o}_l \in \mathbb{R}^{D}$ and the human joint tokens $\mathbf{h}_l \in \mathbb{R}^{J \times D}$ to form $\mathbf{y}_l \in \mathbb{R}^{(J+1) \times D}$. The concatenation is then fed into three linear projections to produce queries $\mathbf{Q}_l$, keys $\mathbf{K}_l$, and values $\mathbf{V}_l$. Formally, at the $i$-th module, the output is computed as: 
\begin{equation} 
    \hat{\mathbf{y}_l} = \text{softmax}\left(\frac{\mathbf{Q}\mathbf{K}^T}{\sqrt{D}} + \mathbf{M}_l^i \right)\mathbf{V}, 
\end{equation} 
where $\mathbf{M}_l^i$ is the dynamic interaction mask. Visible joints are set to 0 in the mask, while masked joints receive $-\infty$.

\begin{figure}
    \centering
    \includegraphics[width=0.9\linewidth]{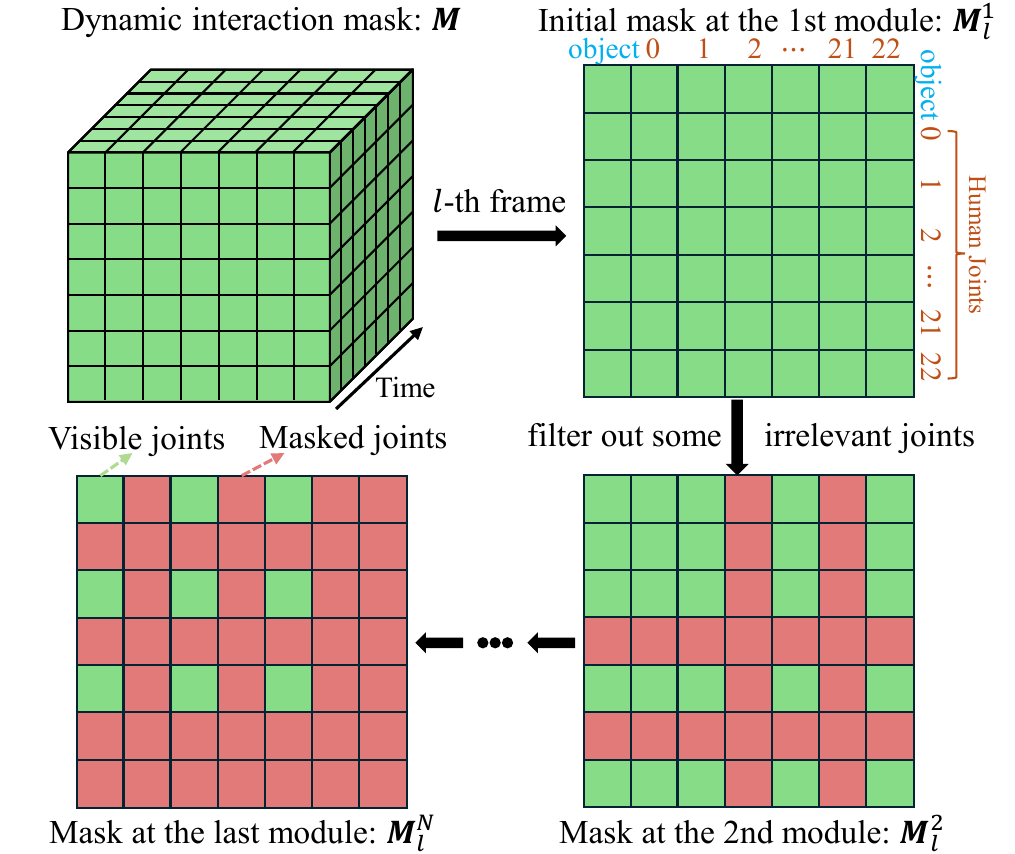}
    \vspace{-0.3cm}
    \caption{Our progressive masking mechanism. Initially, all joints are visible. At each Joint-level Interaction Module, we filter out $k$ joints with the lowest attention scores, leading to more accurate interaction modeling.}
    \vspace{-0.5cm}
    \label{fig:mask}
\end{figure}

% \vspace{0.1cm}
\noindent\textbf{Progressive Masking Mechanism.} Since masking all irrelevant joints at once may lead to errors, we propose a progressive masking mechanism that masks only a few joints at a time, as shown in \cref{fig:mask}. We use the attention scores between the object and joints from the second DIB as a metric to assess their interaction potential. Because the features are more informative after the DCI, at each step, we mask the $k$ joints with the lowest scores. Thus, the DIB progressively refines the joint selection, achieving more accurate interaction modeling.

\subsection{Training Losses}\label{sec:loss}
During training, we adopt three losses, \ie, diffusion loss, object loss, and smooth loss, to optimize our EJIM:
\begin{equation}
    \mathcal{L} = \lambda_1 ||\mathbf{x} - \hat{\mathbf{x}}||_2 + \lambda_2 || \mathbf{x}^o - \hat{\mathbf{x}}^o||_2  + \lambda_3 \sum^{L-1}_{l=1} || \mathbf{x}_l - \mathbf{x}_{l+1}||_2,
\end{equation}
where $\mathbf{x} = \{\mathbf{x}^h, \mathbf{x}^o\}$ denotes the generated HOI sequence, $\hat{\mathbf{x}} = \{\hat{\mathbf{x}}^h, \hat{\mathbf{x}}^o\}$ denotes the ground-truth HOI sequence. The object loss directly constrains the object’s position, while the smooth loss mitigates jitter. We set $\lambda_1 = \lambda_2 = 1$ and $\lambda_3 = 0.5$.

\begin{table*}[h]
\caption{Quantitative results on the BEHAVE \cite{behave} and OMOMO \cite{omomo} test sets. Each evaluation was conducted 20 times to compute  average results with a 95\% confidence interval (denoted as ±). The best performance is in bold, and the second-best is underlined. Average Inference Time (AIT), calculated only on the BEHAVE dataset, denotes the mean over 100 samples on an RTX 3090.}
\vspace{-0.3cm}

\centering
\setlength\tabcolsep{1mm}
\begin{tabular}{lccccccccc}
\bottomrule
 \multirow{2}{*}{Methods} & \multirow{2}{*}{AIT$\downarrow$} & \multirow{2}{*}{FID$\downarrow$}  & \multicolumn{3}{c}{R-Precision$\uparrow$} & \multirow{2}{*}{MM Dist $\downarrow$} & \multirow{2}{*}{CD $\downarrow$} & \multirow{2}{*}{PS $\downarrow$} & \multirow{2}{*}{FSR $\downarrow$}
 \\ \cline{4-6}
   & & & Top1 & Top2 & Top3 & & \\ \toprule
\rowcolor{gray!30} \multicolumn{10}{l}{\textit{On the BEHAVE dataset}}\\ \hline 
% Ground-truth & - & 0 & $0.287^{\pm.011}$ & $0.443^{\pm.013}$ & $0.544^{\pm.011}$ & - & - & -\\
MDM$^{finetuned}$ \cite{mdm} & $\underline{5.31}$s  & $0.246^{\pm.006}$ & $0.223^{\pm.011}$ & $0.378^{\pm.105}$ & $0.488^{\pm.015}$ &$1.118^{\pm.006}$ & - & - & -  \\
MDM$^\star$ \cite{mdm} &$5.34$s & $0.257^{\pm.004}$ & $0.220^{\pm.007}$ & $0.355^{\pm.001}$ & $0.451^{\pm.001}$ & $1.071^{\pm.003}$ & $0.481^{\pm.014}$ & $0.095^{\pm.007}$ & $0.098^{\pm.002}$\\
PriorMDM$^\star$ \cite{priorMDM} & $38.4$s & $0.328^{\pm.018}$ & $0.243^{\pm.009}$ & $0.329^{\pm.009}$ & $0.385^{\pm.013}$  & $1.203^{\pm.012}$ & $0.232^{\pm.010}$ & $0.116^{\pm.001}$ & $\underline{0.066}^{\pm.004}$\\
InerDiff \cite{interdiff} & $16.5$s & $0.170^{\pm.002}$ & $0.310^{\pm.003}$ & $0.480^{\pm.005}$ & $0.599^{\pm.001}$  & $\underline{1.045}^{\pm.002}$ & $0.206^{\pm.003}$ & $\pmb{0.078}^{\pm.000}$ & $0.069^{\pm.002}$\\
CHOIS$^\star$ \cite{chois} & $5.80$s & $\underline{0.157}^{\pm.001}$ & $0.301^{\pm.002}$ & $\underline{0.488}^{\pm.003}$ & $\underline{0.606}^{\pm.003}$  & $1.090^{\pm.002}$ & $0.202^{\pm.003}$ & $0.086^{\pm.001}$ & $0.118^{\pm.003}$ \\
HOI-Diff \cite{hoidiff} & $5.97$s & $0.437^{\pm.004}$ & $\underline{0.312}^{\pm.002}$ & $0.467^{\pm.003}$ & $0.563^{\pm.006}$  &$1.107^{\pm.003}$ & $\underline{0.117}^{\pm.003}$ & $\underline{0.081}^{\pm.001}$ & $0.098^{\pm.002}$ \\ \hline
\textbf{Our EJIM } & {$\pmb{0.27}$s} & $\pmb{0.124}^{\pm.001}$ & $\pmb{0.403}^{\pm.010}$ & $\pmb{0.583}^{\pm.010}$ & $\pmb{0.693}^{\pm.009}$  & $\pmb{0.983}^{\pm.003}$ & $\pmb{0.107}^{\pm.002}$ & $0.083^{\pm.001}$ & $\pmb{0.057}^{\pm.001}$\\ \toprule
\rowcolor{gray!30} \multicolumn{10}{l}{\textit{On the OMOMO dataset}}\\ \hline 
% Ground-truth & - & 0 & $0.247^{\pm.006}$ & $0.398^{\pm.004}$ & $0.504^{\pm.005}$ & - & - & -\\
MDM$^{finetuned}$ \cite{mdm} &  & $\underline{0.164}^{\pm.004}$ & $0.123^{\pm.006}$ & $0.208^{\pm.006}$ & $0.278^{\pm.007}$  & $1.228^{\pm.004}$ & - & - & - \\
MDM$^\star$ \cite{mdm} &  & $0.169^{\pm.005}$ & $0.120^{\pm.004}$ & $0.208^{\pm.006}$ & $0.281^{\pm.009}$ & $1.191^{\pm.004}$ & $0.686^{\pm.002}$ & $0.022^{\pm.006}$ & $0.134^{\pm.001}$\\
PriorMDM$^\star$ \cite{priorMDM}&  & $0.329^{\pm.001}$ & $\underline{0.147}^{\pm.004}$ & $0.219^{\pm.007}$ & $0.277^{\pm.005}$  & $1.200^{\pm.005}$ & $0.755^{\pm.022}$ & $0.025^{\pm.001}$ & $\underline{0.115}^{\pm.007}$\\
InerDiff \cite{interdiff}&  & $0.253^{\pm.007}$ & $0.118^{\pm.009}$ & $0.210^{\pm.009}$ & $0.281^{\pm.007}$  & $\underline{1.167}^{\pm.001}$ & $0.585^{\pm.003}$ & $\pmb{0.015}^{\pm.001}$ & $0.139^{\pm.001}$\\
CHOIS$^\star$ \cite{chois}&  & $0.251^{\pm.013}$ & $0.133^{\pm.003}$ & $\underline{0.254}^{\pm.002}$ & $\underline{0.343}^{\pm.003}$  & $1.193^{\pm.003}$ & $0.433^{\pm.001}$ & $0.021^{\pm.001}$ & $0.151^{\pm.004}$ \\
HOI-Diff \cite{hoidiff}&  & $0.245^{\pm.001}$ & $0.140^{\pm.002}$ & $0.253^{\pm.004}$ & $0.340^{\pm.001}$  & $1.183^{\pm.005}$ & $\underline{0.331}^{\pm.015}$ & $\underline{0.017}^{\pm.001}$ & $0.136^{\pm.004}$\\ \hline
\textbf{Our EJIM} & & $\pmb{0.127}^{\pm.002}$ & $\pmb{0.194}^{\pm.005}$ & $\pmb{0.322}^{\pm.006}$ & $\pmb{0.429}^{\pm.002}$  & $\pmb{1.130}^{\pm.004}$ & $\pmb{0.300}^{\pm.009}$ & $0.020^{\pm.001}$ & $\pmb{0.111}^{\pm.001}$\\ \toprule
\end{tabular}
\label{tab:main}
\vspace{-0.4cm}
\end{table*}

\section{Experiments}
\subsection{Experimentation Details}

\noindent\textbf{Datasets.}
We conduct experiments on two common public datasets: BEHAVE \cite{behave} and OMOMO \cite{omomo}. BEHAVE has 20 objects, 8 subjects, and 1,451 sequences, and we use its HOI-Diff \cite{hoidiff} preprocessed version with text descriptions, following the official train-test split. OMOMO comprises 10 hours of sequences with 15 objects and 17 subjects, also with text descriptions. We apply the same HOI-Diff-based preprocessing and ensure that objects differ between training and testing sets.

% \vspace{0.1cm}
% \noindent\textbf{Implementation Details.}
% We set the maximum diffusion step $T$ to 1000, with variances $\beta_t$ ranging from $1\times10^{-4}$ to $2\times10^{-2}$. We use DDIM \cite{DDIM} with 50 steps during sampling. The latent dimension $D$ and the number of modules $N$ are set to 128 and 6, respectively.  Each submodule in DHM and DCI is implemented as two Mamba blocks \cite{mamba}. We train EJIM with AdamW \cite{adamw}, using a learning rate of $10^{-4}$, a batch size of 32, and 150 epochs on two RTX 3090Ti GPUs.

% \vspace{0.1cm}
\noindent\textbf{Evaluation Metrics.}
Following \cite{light-t2m,hoidiff,pmg}, we adopt the metrics from \cite{t2m}: Frechet Inception Distance (FID), R-Precision, and MultiModal Distance (MM Dist). FID measures distribution similarity between generated and ground-truth motions in latent space, R-Precision evaluates text-motion consistency, and MM Dist assesses the feature distance between given texts and generated motions. Since these metrics require a trained evaluator and no public one is available, we train a evaluator via contrastive learning, following \cite{cghoi, hoidiff}. 

We further assess human-object interaction using contact distance (CD), penetration score (PS), and foot skating rate (FSR). CD calculates the distance between the interaction joint and object mesh, while the interaction joint and contact label are obtained from GT. PS is the proportion of the human mesh with negative SDF relative to the object mesh, and FSR quantifies foot sliding \cite{chois}. Refer to Appendix for more details. 

\subsection{Comparisons to Existing Methods}

\noindent\textbf{Compared Methods.}
Due to inconsistencies in data representation and text annotations in previous methods, we reproduce the available open-source methods for text-to-motion or HOI generation.  % Since our HOI representation can be converted to the format used by HOI-Diff, we directly evaluate HOI-Diff’s released checkpoints using our metrics. 
A brief implementation of compared methods is as follows (details are shown in Appendix): (a) MDM$^{finetuned}$ \cite{mdm}:  As MDM \cite{mdm} is a text-driven motion method, we directly fine-tune the pretrained MDM model on HOI datasets that focus solely on human motion generation. (b) MDM$^\star$ \cite{mdm}:  We concatenate human motion and object motion along the time dimension, then train MDM from scratch to produce HOI sequences. (c) PriorMDM$^\star$ \cite{priorMDM}: Adapted from a two-person motion framework, we replace one human with an object to generate HOI sequences. (d) InterDiff \cite{interdiff}: Since InterDiff isn’t originally text-driven, we add text inputs to condition the model for HOI generation.  (e) CHOIS$^\star$ \cite{chois}: As CHOIS uses text and object waypoints to generate HOI, we remove the waypoints and adjust the input/output dimensions.

\begin{table}[t]
    \vspace{-0.3cm}
    \caption{User Study Results. Bold values indicate better performance.}
    \vspace{-0.3cm}
    \centering
    \setlength{\tabcolsep}{0.15cm}
    \resizebox{\linewidth}{!}{
    \begin{tabular}{lccc}
    \toprule
          & InterDiff vs Ours & CHOIS$^\star$ vs Ours & HOI-Diff vs Ours \\ 
    \midrule
         Semantic Matching       & $43.6\%$ vs $\pmb{56.4\%}$ & $47.3\%$ vs $\pmb{52.7\%}$ & $40.0\%$ vs $\pmb{60.0\%}$ \\
         Interaction Plausibility  & $43.6\%$ vs $\pmb{56.4\%}$ & $43.6\%$ vs $\pmb{56.4\%}$ & $47.3\%$ vs $\pmb{52.7\%}$ \\ 
    \bottomrule
    \end{tabular}
    }
     \vspace{-0.3cm}
    \label{tab:user_study}
\end{table}

\begin{table}[t]
    \caption{Ablation study of the main modules on the BEHAVE dataset.} 
    \vspace{-0.3cm}
    \centering
    \resizebox{0.8\linewidth}{!}{
    \begin{tabular}{c ccccc}
    \toprule
        & FID$\downarrow$ & R-Top1$\uparrow$ & CD$\downarrow$ & PS$\downarrow$ \\ \hline
         Single-branch & $0.159$ & $0.307$ & $0.137$ & $\underline{0.080}$ \\ \hline
         w/o DHM & $\underline{0.151}$ & $\underline{0.399}$ & $\underline{0.117}$ & $0.087$\\ 
         w/o DCI & $\pmb{0.124}$ & $0.381$ & $0.139$ & $\pmb{0.079}$\\
         w/o DIB & $0.190$ & $0.357$ & $0.150$ & $0.085$\\ \hline
        \rowcolor{gray!30} Full EJIM & $\pmb{0.124}$ & $\pmb{0.403}$ & $\pmb{0.107}$ & $0.083$ \\
    \toprule
    \end{tabular}
    }
    \vspace{-0.5cm}
    \label{tab:ablation}
\end{table}

\vspace{0.1cm}
\noindent\textbf{Quantitative Comparisons.}
As shown in \cref{tab:main},  EJIM achieves state-of-the-art results on both datasets, demonstrating substantial improvements in human motion quality and interaction quality. Notably, EJIM requires only 4.5\% of the inference time used by HOI-Diff, confirming its efficiency. The large performance gap between EJIM and previous methods highlights the importance of explicit joint-level interaction modeling. Note that a lower PS does not necessarily indicate strong performance when both FID and CD scores are poor.

\begin{figure*}[ht]
    \centering
    \setlength{\tabcolsep}{0cm}
    \begin{tabular}{cccc}
    \toprule
    \makebox[0.25\textwidth][c]{\textbf{Our EJIM}} & \makebox[0.25\textwidth][c]{InterDiff} & \makebox[0.25\textwidth][c]{CHOIS$^\star$} & \makebox[0.23\textwidth][c]{HOI-Diff} \\ \hline
    \multicolumn{4}{c}{An individual is currently \textbf{sitting on the chairwood}.}\vspace{-0.06cm} \\
    \multicolumn{4}{c}{\includegraphics[width=1.0\linewidth]{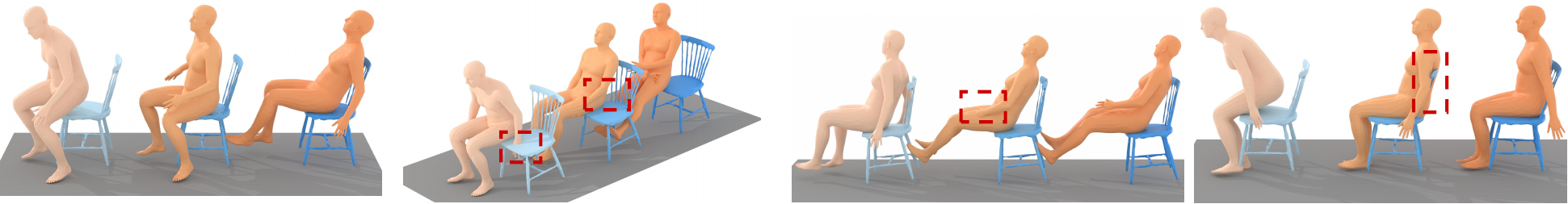}} \vspace{-0.11cm }\\ \hline
    \multicolumn{4}{c}{Someone is applying force to the \textbf{tablesquare} by \textbf{pulling} it \textbf{on the ground}.}\vspace{-0.06cm}\\
    \multicolumn{4}{c}{\includegraphics[width=\linewidth]{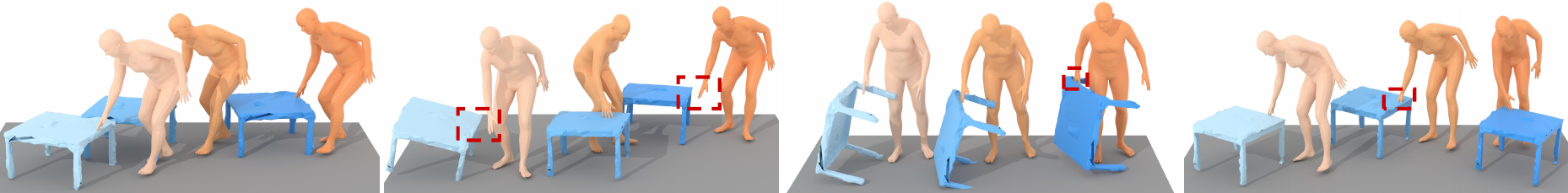}} \vspace{-0.11cm }\\ \hline
    \multicolumn{4}{c}{A person \textbf{moves} the \textbf{stool} by exerting force back and forth \textbf{on the floor}.} \vspace{-0.06cm}\\
    \multicolumn{4}{c}{\includegraphics[width=\linewidth]{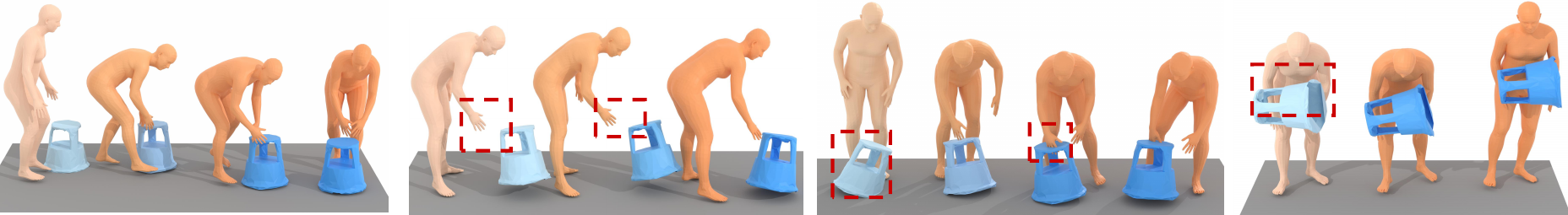}} \vspace{-0.11cm }\\ 
    \toprule
    \end{tabular}
    \vspace{-0.45cm}
    \caption{Qualitative comparisons on the BEHAVE dataset. Red boxes highlight issues like mesh penetration, large contact distances, or text inconsistencies. Our approach generates more realistic and plausible human-object interactions. The mesh color darkens over time to represent progress.}
    \label{fig:vis}
     \vspace{-0.5cm}
\end{figure*}

\vspace{0.1cm}
\noindent\textbf{Qualitative Comparisons.}
As shown in \cref{fig:vis}, InterDiff and CHOIS$^\star$ generate motions with unrealistic contact and significant mesh penetration, while HOI-Diff shows improvements but still struggles with penetration and poor text alignment in some cases. In contrast, EJIM generates smoother, more realistic interactions, accurately capturing spatial relationships and aligning well with textual prompts, outperforming baselines. Video results are provided in the supplementary materials.

\vspace{0.1cm}
\noindent\textbf{User Study.}
Furthermore, we conduct a user study involving 55 users to demonstrate EJIM's superiority. We compare the EJIM model with existing methods one by one and ask users to select preferred animations along two aspects: (1) Semantic Matching: Which one better aligns with the semantic content of the given text descriptions? (2) Interaction Plausibility: Which one displays more realistic and coherent interactions between humans and objects? As shown in \cref{tab:user_study}, results demonstrate that our EJIM outperforms others in both aspects.

\begin{table}[t]
    \vspace{-0.3cm}
    \caption{Ablation study of the DHM on the BEHAVE dataset.}
    \centering
    \vspace{-0.3cm}
    \resizebox{0.95\linewidth}{!}{
    \begin{tabular}{c ccccc}
    \toprule
        & FID$\downarrow$ & R-Top1$\uparrow$ & CD$\downarrow$ & PS$\downarrow$ \\ \hline
        w/o Limb-guided Mamba& $\underline{0.130}$ & $0.396$ & $0.133$ & $0.085$\\ \hline
        Vanilla Scan & $0.143$ & $0.357$ & $\underline{0.123}$ & $\pmb{0.080}$\\
        w/o Learnable Token & $0.141$ & $0.375$ & $0.128$ & $0.083$\\
        Fixed token & $0.131$ & $\pmb{0.419}$ & $0.188$ & $\underline{0.081}$\\ \hline
        \rowcolor{gray!30} Learnable Token (Ours) & $\pmb{0.124}$ & $\underline{0.403}$ & $\pmb{0.107}$ & $0.083$ \\
    \toprule
    \end{tabular}
    }
    \vspace{-0.6cm}
    \label{tab:ablation_DHM}
\end{table}

\vspace{-0.4cm}
\subsection{Ablation Studies} \label{sec:ablation}
\noindent\textbf{Effectiveness of Main Components.}
As shown in \cref{tab:ablation}, we evaluate the effectiveness of each proposed component by individually removing them. Eliminating any component significantly degrades at least one performance metric. Moreover, to validate the necessity of the dual-branch design in our DHM and DCI, we also present results obtained using a single-branch configuration in which human and object motions are concatenated. The results indicate that using a single branch causes substantial declines in FID, R-Top1, and CD.

\noindent\textbf{Analysis of DHM.}
As shown in \cref{tab:ablation_DHM}, we provide a detailed analysis of DHM. Removing the Limb-guided Mamba or replacing it with a Vanilla Scan (as shown in \cref{fig:lgm}.d) results in a significant performance drop. Furthermore, removing learnable tokens that delineates distinct limbs or replacing them with fixed tokens severely degrades FID and CD. Results highlight the importance of each design within our DHM.

\noindent\textbf{Analysis of DIB.}
As shown in \cref{tab:ablation_DIB}, we analyze the impact of our progressive masking mechanism. Disabling the masking mechanism (denoted as k=0) leads to poorer performance on all metrics, confirming its necessity. In addition, we vary the number of masked human joints (k). The results show that performance deteriorates when k is set too low or too high, likely due to redundant information or excessive masking.

\begin{table}[t]
    \vspace{-0.1cm}
    \caption{Ablation study of the number of masked joints (k) in the DIB’s progressive masking mechanism.}
    \vspace{-0.3cm}
    \centering
    \resizebox{0.72\linewidth}{1.25cm}{
    \begin{tabular}{c ccccc}
    \toprule
        & FID$\downarrow$ & R-Top1$\uparrow$ & CD$\downarrow$ & PS$\downarrow$ \\ \hline
         k=0 & $0.154$ & $0.359$ & $0.132$ & $0.088$\\
         k=1 & $0.130$ & $0.359$ & $0.116$ & $0.088$\\
         k=2 & $\underline{0.128}$ & $\underline{0.399}$ & $\pmb{0.104}$ & $\underline{0.085}$\\ 
        \rowcolor{gray!30} k=3 & $\pmb{0.124}$ & $\pmb{0.403}$ & $\underline{0.107}$ & $\pmb{0.083}$ \\
        k=4 & $0.130$ & $0.370$ & $0.120$ & $0.086$\\ 
        
    \toprule
    \end{tabular}
    }
     \vspace{-0.6cm}
    \label{tab:ablation_DIB}
\end{table}

\vspace{-0.2cm}
\section{Conclusions}
In conclusion, our proposed EJIM effectively addresses the challenges of explicit joint-level HOI modeling by integrating advanced modules for spatiotemporal information processing, semantic conditioning, and progressive joint mask. This results in significantly more accurate and realistic HOIs, while achieving remarkable computational efficiency.

\noindent\textbf{Acknowledgments.}
This work was supported partially by NSFC(92470202, U21A20471), National Key Research and Development Program of China (2023YFA1008503), Guangdong NSF Project (No. 2023B1515040025).

\vspace{-0.17cm}
\bibliographystyle{IEEEbib}
\bibliography{icme2025references}

\clearpage

\appendices

\twocolumn[{%  <-- 注意这里使用了 \twocolumn[{ ... }]
\begin{center}
  \Large\bfseries Appendix of ``Efficient Explicit Joint-level Interaction Modeling with Mamba for Text-guided HOI Generation'' % 使用 \Large 和 \bfseries 设置标题格式
  \vspace{1em} % 可选：添加一些垂直间距
\end{center}
}] % <--  \twocolumn 的参数结束

\section{Summary}
In the appendix, we first introduce the new HOI representation format in \cref{scheme} in detail. Next, in \cref{experiment}, we provide additional details about the experimental setup, including training implementation details (\cref{implementation}), explanations of evaluation metrics (\cref{metrics}), specifics of our evaluator (\cref{evaluator}), and detailed configurations of the compared methods (\cref{compared}). Then, in \cref{ablation}, we conduct extensive ablation studies to evaluate the impact of various design choices and hyperparameters. Finally, in \cref{limitations}, we discuss the limitations of our EJIM.

\section{Details of the new HOI representation}\label{scheme}
In this section, we provide a detailed explanation of the novel human-object interaction (HOI) representation mentioned in our main manuscript (Sec. III-B), which preserves the information for each joint. Specifically, each non-root joint is represented as a 12-dimensional vector that includes the local joint position $j^p \in \mathbb{R}^3$,
local joint velocity $j^v \in \mathbb{R}^3$, and joint rotation $j^r \in \mathbb{R}^6$  in the root space. The root joint is represented as a 7-dimensional vector, consisting of the global position $r^p \in \mathbb{R}^3$,
angular velocity along the Y-axis $r^a \in \mathbb{R}$, linear velocity on the XZ-plane $r^v \in \mathbb{R}^2$, and the height $r^y \in \mathbb{R}$. The object node is represented by a 6-dimensional vector, which captures its global rotation and translation information. Additionally, to mitigate foot skating, we introduce a virtual joint represented by a 4-dimensional binary vector  $c^f \in \mathbb{R}^4$, indicating foot-ground contact features. Finally, to ensure uniformity, the representations of the virtual joint and object node are zero-padded to match the
$D_{in}$ dimensions ($D_{in}=12$). This representation allows us to efficiently preserve the motion information for each joint and object node, providing a comprehensive description of the 3D HOI sequence.

\section{More Details of the Experiment}\label{experiment}
In this section, we outline the experimental setup, including implementation details, evaluation metrics, the evaluator's design, and the details of other comparative methods.

\subsection{Implementation Details.}\label{implementation}
We set the maximum diffusion step $T$ to 1000, with variances $\beta_t$ ranging from $1\times10^{-4}$ to $2\times10^{-2}$. We use DDIM \cite{DDIM} with 50 steps during sampling. The latent dimension $D$ and the number of modules $N$ are set to 128 and 6, respectively.  Each submodule in DHM and DCI is implemented as two Mamba blocks \cite{mamba}. We train EJIM with AdamW \cite{adamw}, using a learning rate of $10^{-4}$, a batch size of 32, and 150 epochs on two RTX 3090Ti GPUs. During testing, the guidance scale is set to 2.

\begin{table*}[ht]
\caption{\textbf{Quantitative evaluation of the BEHAVE \cite{behave} and OMOMO \cite{omomo} test sets.} We repeated evaluation 20 times to calculate the average results with a 95\% confidence interval (denoted by ±). The best result is in bold, and the second best is underlined.}
\centering
\setlength\tabcolsep{1mm}
\resizebox{\linewidth}{!}{
\begin{tabular}{lccccccccc}
\bottomrule
 \multirow{2}{*}{Methods} & \multirow{2}{*}{FID$\downarrow$}  & \multicolumn{3}{c}{R-Precision$\uparrow$} & \multirow{2}{*}{MM Dist$\downarrow$} & \multirow{2}{*}{Div.$\uparrow$} & \multirow{2}{*}{CD $\downarrow$} & \multirow{2}{*}{PS $\downarrow$} & \multirow{2}{*}{FSR $\downarrow$}\\ \cline{3-5}
   & & Top1 & Top2 & Top3 & & \\ \toprule
\rowcolor{gray!30} \multicolumn{10}{l}{\textit{On the BEHAVE dataset}}\\ \hline 
Real motion. & $0.001^{\pm.000}$ & $0.287^{\pm.011}$ & $0.443^{\pm.013}$ & $0.544^{\pm.011}$ & $1.055^{\pm.002}$ & $1.346^{\pm.011}$ & - & - & -\\ \hline
MDM$^{finetuned}$ \cite{mdm} & $0.246^{\pm.006}$ & $0.223^{\pm.011}$ & $0.378^{\pm.105}$ & $0.488^{\pm.015}$ & $1.118^{\pm.006}$ & $\underline{1.350}^{\pm.008}$ & - & - & - \\
MDM$^\star$ \cite{mdm} & $0.257^{\pm.004}$ & $0.220^{\pm.007}$ & $0.355^{\pm.001}$ & $0.451^{\pm.001}$ & $1.071^{\pm.003}$ & $1.307^{\pm.006}$ & $0.481^{\pm.014}$ & $0.095^{\pm.007}$ & $0.098^{\pm.002}$ \\
PriorMDM$^\star$ \cite{priorMDM}& $0.328^{\pm.018}$ & $0.243^{\pm.009}$ & $0.329^{\pm.009}$ & $0.385^{\pm.013}$ & $1.203^{\pm.012}$ & $1.142^{\pm.017}$ & $0.232^{\pm.010}$ & $0.116^{\pm.001}$ & $\underline{0.066}^{\pm.004}$\\
InerDiff \cite{interdiff}  & $0.170^{\pm.002}$ & $0.310^{\pm.003}$ & $0.480^{\pm.005}$ & $0.599^{\pm.001}$ & $\underline{1.045}^{\pm.002}$ & $1.325^{\pm.006}$ & $0.206^{\pm.003}$ & $\pmb{0.078}^{\pm.000}$ & $0.069^{\pm.002}$\\
CHOIS$^\star$ \cite{chois}  & $\underline{0.157}^{\pm.001}$ & $0.301^{\pm.002}$ & $\underline{0.488}^{\pm.003}$ & $\underline{0.606}^{\pm.003}$ & $1.090^{\pm.002}$ & $1.265^{\pm.013}$ & $0.202^{\pm.003}$ & $0.086^{\pm.001}$ & $0.118^{\pm.003}$\\
HOI-Diff \cite{hoidiff} & $0.437^{\pm.004}$ & $\underline{0.312}^{\pm.002}$ & $0.467^{\pm.003}$ & $0.563^{\pm.006}$ & $1.107^{\pm.003}$ & $1.235^{\pm.020}$ & $\underline{0.117}^{\pm.003}$ & $\underline{0.081}^{\pm.001}$ & $0.098^{\pm.002}$\\ \hline
\textbf{Our EJIM }  & $\pmb{0.124}^{\pm.001}$ & $\pmb{0.403}^{\pm.010}$ & $\pmb{0.583}^{\pm.010}$ & $\pmb{0.693}^{\pm.009}$ & $\pmb{0.983}^{\pm.003}$ & $\pmb{1.365}^{\pm.018}$ & $\pmb{0.107}^{\pm.002}$ & $0.083^{\pm.001}$ & $\pmb{0.057}^{\pm.001}$\\ \toprule
\rowcolor{gray!30} \multicolumn{10}{l}{\textit{On the OMOMO dataset}}\\ \hline 
Real motion.  & $0.001^{\pm.001}$ & $0.247^{\pm.006}$ & $0.398^{\pm.004}$ & $0.504^{\pm.005}$ & $1.050^{\pm.001}$ & $1.356^{\pm.013}$ & - & - & -\\ \hline
MDM$^{finetuned}$ \cite{mdm}   & $\underline{0.164}^{\pm.004}$ & $0.123^{\pm.006}$ & $0.208^{\pm.006}$ & $0.278^{\pm.007}$ & $1.228^{\pm.004}$ & $\underline{1.333}^{\pm.002}$ & - & - & -\\
MDM$^\star$ \cite{mdm}   & $0.169^{\pm.005}$ & $0.120^{\pm.004}$ & $0.208^{\pm.006}$ & $0.281^{\pm.009}$ & $1.191^{\pm.004}$ & $1.319^{\pm.001}$ & $0.686^{\pm.002}$ & $0.022^{\pm.006}$ & $0.134^{\pm.001}$ \\
PriorMDM$^\star$ \cite{priorMDM} & $0.329^{\pm.001}$ & $\underline{0.147}^{\pm.004}$ & $0.219^{\pm.007}$ & $0.277^{\pm.005}$ & $1.200^{\pm.005}$ & $1.181^{\pm.003}$ & $0.755^{\pm.022}$ & $0.025^{\pm.001}$ & $\underline{0.115}^{\pm.007}$ \\
InerDiff \cite{interdiff} & $0.253^{\pm.007}$ & $0.118^{\pm.009}$ & $0.210^{\pm.009}$ & $0.281^{\pm.007}$ & $\underline{1.167}^{\pm.001}$ & $1.227^{\pm.003}$ & $0.585^{\pm.003}$ & $\pmb{0.015}^{\pm.001}$ & $0.139^{\pm.001}$ \\
CHOIS$^\star$ \cite{chois} & $0.251^{\pm.013}$ & $0.133^{\pm.003}$ & $\underline{0.254}^{\pm.002}$ & $\underline{0.343}^{\pm.003}$ & $1.193^{\pm.003}$ & $\pmb{1.334}^{\pm.014}$ & $0.433^{\pm.001}$ & $0.021^{\pm.001}$ & $0.151^{\pm.004}$ \\
HOI-Diff\cite{hoidiff} & $0.245^{\pm.001}$ & $0.140^{\pm.002}$ & $0.253^{\pm.004}$ & $0.340^{\pm.001}$ & $1.183^{\pm.005}$ & $1.303^{\pm.014}$ & $\underline{0.331}^{\pm.015}$ & $\underline{0.017}^{\pm.001}$ & $0.136^{\pm.004}$ \\ \hline
\textbf{Our EJIM}  & $\pmb{0.127}^{\pm.002}$ & $\pmb{0.194}^{\pm.005}$ & $\pmb{0.322}^{\pm.006}$ & $\pmb{0.429}^{\pm.002}$ & $\pmb{1.130}^{\pm.004}$ & $1.327^{\pm.017}$ & $\pmb{0.300}^{\pm.009}$ & $0.020^{\pm.001}$ & $\pmb{0.111}^{\pm.001}$ \\ \toprule
\end{tabular}
}
\label{tab:main}
\end{table*}

\subsection{Evaluation Metrics.}\label{metrics}
In this section, we specifically introduce all the evaluation metrics used, focusing on motion generation evaluation and interaction quality assessment.

\noindent\textbf{Motion Generation Evaluation.}
We adopt Frechet Inception Distance (FID), R-Precision, MultiModal Distance (MM Dist), and Diversity (Div.) from \cite{t2m} to evaluate the quality of generated motions.

\begin{itemize}
\item \textit{Frechet Inception Distance} (FID): FID measures the similarity between the distributions of generated and real motions in the latent feature space. A lower FID score indicates that the quality of the generated motions is closer to that of the real motions.
\item \textit{R-Precision}: This metric evaluates the semantic consistency between generated motions and text descriptions by calculating the proportion of correctly matched text descriptions.
\item \textit{MultiModal Distance} (MM Dist): MM Dist quantifies the feature distance between the given text and the generated motions. A smaller MM Dist value suggests better semantic alignment between the generated motions and the text descriptions.
\item \textit{Diversity} (Div.): We use an additional metric, Div., to further evaluate the variety of generated motions. Specifically, we compute the average distance between two randomly sampled equal-sized subsets of the generated motions to evaluate the diversity,
\end{itemize}

\noindent\textbf{Human-object Interaction Evaluation.}
To better evaluate the quality of the model's generation, we also introduce Contact Distance (CD), Penetration Score (PS), and Foot Skating Rate (FSR) to assess human-object interaction.
\begin{itemize}
\item \textit{Contact Distance} (CD): CD quantifies the distance between human joints and the object mesh, representing the precision of human-object interaction. The positions of the interaction joints and the contact labels are derived from ground truth (GT) data. Lower CD values indicate more precise interactions
\item \textit{Penetration Score} (PS): PS measures the proportion of human body mesh vertices with negative Signed Distance Function (SDF) values relative to the object mesh, indicating the degree of penetration. Lower PS values suggest more physically plausible motions with fewer unnatural interaction errors.
\item \textit{Foot Skating Rate} (FSR): FSR measures foot sliding during motion generation, calculated as the proportion of frames where foot height is below 5 cm and sliding distance exceeds 2.5 cm. Lower FSR values indicate more natural, physically accurate motions.
\end{itemize}
As shown in \cref{tab:main}, the table extends the main table in our manuscript by including Real motion results and the Div. metric across all methods. The results show that EJIM excels in FID for motion fidelity, R-Precision for semantic alignment, and achieves the best CD, PS, and FSR scores, indicating precise interactions and reduced foot sliding. By balancing fidelity, diversity, and interaction precision, EJIM outperforms baselines, proving its effectiveness in generating realistic HOI sequences. Furthermore, as reported in studies like \cite{hoianimator, momask, bamm, mmm, zeng,zeng2020hybrid}, 
it is possible for some models to achieve R-Precision scores exceeding those of real motions.

\subsection{Details of Our Evaluator.}\label{evaluator}
As mentioned in the main manuscript, since the motion generation evaluation metrics require a trained evaluator and no public model is currently available, we follow \cite{cghoi, hoidiff} to train an evaluator using contrastive learning.

\noindent\textbf{Overview of Our Evaluator.} 
To evaluate the quality of motion generation, we propose a custom evaluator inspired by CLIP \cite{clip}, based on a contrastive learning framework. This evaluator aligns the semantic representations of motion and text through two modules. One module processes textual descriptions by appending a CLS token and then applying a linear projection to generate feature representations. The other module handles motion sequences in a similar way by appending a CLS token and applying linear projection. Finally, the evaluator computes the contrastive loss to effectively align motion and text representations within a shared semantic space. 
Specifically, the text module uses the pre-trained RoBERTa model \cite{roberta} to generate 512-dimensional textual embeddings. The motion module, designed to process human motion data in the HumanML3D format \cite{t2m} with a feature dimension of 263, consists of 8 Transformer decoders \cite{transformer} with a hidden dimension of 384 and outputs 512-dimensional embeddings.

\subsection{Compared Methods.}\label{compared}
We compare our approach with various related methods on the BEHAVE\cite{behave} and OMOMO\cite{omomo} datasets. Due to inconsistencies in data representation and text annotations in some methods, we adjust the original implementations to make them suitable for HOI generation tasks. Further details are provided below.

\begin{itemize}
    \item MDM$^{finetuned}$ \cite{mdm}: MDM is a text-driven human motion generation method that ignores object interactions. We fine-tune the pretrained MDM model (MDM$^{finetuned}$) on our HOI datasets to generate human motion only, without producing object dynamics or evaluating interaction quality metrics.
    \item  MDM$^\star$ \cite{mdm}: To adapt MDM for text-driven HOI generation, we modify its input and output by concatenating human motion and object 6-DoF motion along the time dimension. This enables MDM$^\star$ to jointly learn human and object motion, generating complete HOI sequences. The model is trained from scratch.
    \item PriorMDM$^\star$ \cite{priorMDM}: While PriorMDM is originally designed for two-person motion generation using the ComMDM module to model interactions, PriorMDM$^\star$ replaces one person with an object and adjusts the input and output dimensions for text-driven HOI generation, enabling it to generate complete human-object interaction sequences.
    \item InterDiff \cite{interdiff}: InterDiff is originally designed to predict future HOI sequences from past ones and does not support text-driven HOI generation. To adapt it, we replace past HOI sequences with text descriptions, add text as a condition, adjust feature dimensions, and use the CLIP text encoder to extract textual features, enabling text-driven HOI generation.
    \item CHOIS$^\star$ \cite{chois}: CHOIS is an HOI generation method that uses text descriptions and object waypoints. To adapt it to our task, we remove its dependency on object waypoints and adjust the input and output dimensions to align with our HOI representation. CHOIS* generates HOI sequences and is fully compatible with our representation.
\end{itemize}

\section{More Ablation Studies}\label{ablation}
In this section, we conduct comprehensive ablation studies to further assess the effectiveness of each component and design choice in our EJIM.

\begin{table}[t]
    \caption{Impact of Weights on the training loss. The gray line indicates the configuration adopted in our EJIM.}
    \centering
    \resizebox{\linewidth}{!}{
    \begin{tabular}{cccccccc}
    \toprule
        $\lambda_2$ & $\lambda_3$ & FID$\downarrow$ & R-Top1$\uparrow$ & CD$\downarrow$ & PS$\downarrow$ & FSR$\downarrow$\\ \hline
         0 & 0 & 0.226 & 0.381 & 0.154& 0.079 &0.086 \\ \hline
         0 & 0.5 & 0.120 & 0.404 & 0.160 & 0.088 &\pmb{0.056} \\
         0.5 & 0.5 & \pmb{0.119} & \pmb{0.422} & 0.193 & 0.084 &0.060 \\
         \rowcolor{gray!30}1 & 0.5 & 0.124 & 0.403 & \pmb{0.107} & \pmb{0.083} &0.057\\
         1.5 & 0.5 & 0.124 & 0.409 & 0.149 & \pmb{0.083} &0.058 \\
         2 & 0.5 &0.152 &0.363 &0.147 &0.088 &0.057 \\ \hline
         
         1 & 0 & 0.162 & 0.359 & 0.132 & 0.087 &0.098 \\
         \rowcolor{gray!30}1 & 0.5 & \pmb{0.124} & \pmb{0.403} &\pmb{0.107} &0.083 &\pmb{0.057} \\ 
         1 & 1 &0.150 &0.386 &0.152 &\pmb{0.081} &0.068 \\
         1 & 1.5 & 0.129 & 0.381 &0.151 & 0.083 &0.061 \\
    \toprule
    \end{tabular}
    }
    \label{tab:loss}
\end{table}

\begin{table}[t]
    \caption{Impact of the guidance scale. The gray line represents the configuration used in our EJIM.}
    \centering
    \resizebox{0.9\linewidth}{!}{
    \begin{tabular}{cccccc}
    \toprule
        \makecell{Guidance \\Scale}  & FID$\downarrow$ & R-Top1$\uparrow$ & CD$\downarrow$ & PS$\downarrow$ \\ \hline
         1 & $0.126$ & $0.293$ & $0.116$ & $0.087$ \\
         \rowcolor{gray!30} 2 & $\pmb{0.124}$ & $0.403$ & $\pmb{0.107}$ & $0.083$\\
         3  & $0.128$ & $0.411$ & $0.139$ & $0.083$ \\
         4 & $0.127$ & $0.434$ & $0.187$ & $0.079$\\
         5 & $0.134$ & $\pmb{0.438}$ & $0.243$ & $\pmb{0.077}$\\
    \toprule
    \end{tabular}
    }

    \label{tab:guidance}
\end{table}

\begin{table}[t]
   \caption{Ablation Study of the Object Geometry on the BEHAVE Dataset}
    \centering
    \setlength{\tabcolsep}{0.1cm}
    \resizebox{0.95\linewidth}{!}{
    \begin{tabular}{c ccccc}
    \toprule
        & FID$\downarrow$ & R-Top1$\uparrow$ & CD$\downarrow$ & PS$\downarrow$ \\ \hline
        w/o object geometry & $0.139$ & $\pmb{0.442}$ & $0.159$ & $0.088$\\
        \rowcolor{gray!30} EJIM & $\pmb{0.124}$ & $0.403$ & $\pmb{0.107}$ & $\pmb{0.083}$ \\
    \toprule
    \end{tabular}
    }
    \label{tab:obj_geo}
\end{table}

\begin{table}[t]
    \caption{Impact of the number of inference steps. The gray line represents the configuration used in our EJIM. The Average Inference Time (AIT) is the mean over 100 samples on an RTX 3090Ti.}
    \centering
    \resizebox{\linewidth}{!}{
    \begin{tabular}{cccccc}
    \toprule
        \makecell{Inference \\Steps} & AIT & FID$\downarrow$ & R-Top1$\uparrow$ & CD$\downarrow$ & PS$\downarrow$ \\ \hline
         20 & \textbf{0.11s} & $0.127$ & $0.358$ & $0.111$ & $0.086$ \\
         \rowcolor{gray!30} 50 & 0.27s& $0.124$ & $0.403$ & $0.107$ & $0.083$\\
         100  & 0.55s & $0.121$ & $0.410$ & $0.107$ & $0.081$ \\
         200 & 1.07s & $\pmb{0.117}$ & $\pmb{0.424}$ & $\pmb{0.104}$ & $\pmb{0.080}$\\
    \toprule
    \end{tabular}
    }

    \label{tab:steps}
\end{table}

\begin{table}[t]
    \caption{Impact of the number of blocks. The gray line represents the configuration used in our EJIM.}
    \centering
    \resizebox{0.9\linewidth}{!}{
    \begin{tabular}{cccccc}
    \toprule
        \#Blocks  & FID$\downarrow$ & R-Top1$\uparrow$ & CD$\downarrow$ & PS$\downarrow$ \\ \hline
         2 & 0.211 & 0.319 & 0.181 & \pmb{0.078}\\
         4 & 0.168 & 0.396 & 0.141 & 0.079\\
         \rowcolor{gray!30} 6 & 0.124 & $\pmb{0.403}$ & $\pmb{0.107}$ &0.083\\
         8 & \pmb{0.123} & 0.398 & 0.170 & 0.086\\
    \toprule
    \end{tabular}
    }
    
    \label{tab:block}
\end{table}
\subsection{Impact of Weights on the Training Loss}
We introduce two loss functions to improve the quality of human-object interaction generation in our EJIM (mentioned in Sec. III-G of the main manuscript). To assess the impact of these loss functions, we evaluate performance by adjusting the weights of each loss.

The evaluation results are presented in \cref{tab:loss}. As shown in the first row, the introduction of object loss and smooth loss significantly improves the generated quality. In the middle part, explicitly constraining the object's 6-DoF substantially enhances performance on CD and PS. In the lower part, the smooth loss used for mitigating jitter significantly improves FID and FSR performance.

\subsection{Impact of the Guidance Scale}
We analyze the impact of the guidance scale on generation performance, as shown in \cref{tab:guidance}. When the guidance scale is set to 1, weak text guidance can cause the generated HOIs to deviate from the input text, resulting in a lower R-Top1 score. However, the FID and CD metrics remain satisfactory, indicating that the model effectively utilizes object geometry to guide HOI generation. As the guidance scale increases, stronger text guidance improves the alignment between the generated HOIs and the text descriptions, leading to a gradual improvement in R-Top1. However, this also introduces inconsistencies in contact distances, likely because stronger text guidance interferes with the accuracy of object geometry modeling, which is crucial for human-object interaction.

\subsection{Impact of the Object Geometry}
We also explore the impact of object geometry on the quality of HOI generation, as shown in \cref{tab:obj_geo}. When object geometry is excluded as a condition, the R-Top1 score improves significantly. However, metrics such as FID, CD, and PS degrade, suggesting a decline in the quality of human-object interaction. We note that the model relies solely on text descriptions, lacking geometric information, which hinders its ability to accurately capture contact relationships and spatial positioning in human-object interactions.

\subsection{Impact of the Inference Steps}
We also evaluate the impact of the number of inference steps. Based on the results in \cref{tab:steps}, as the number of inference steps increases, the model's performance improves across all metrics. However, this improvement in performance comes with an increase in inference time. With 20 inference steps, the Average Inference Time (AIT) is 0.11s, while with 200 steps, the inference time increases 9.7 times, reaching 1.07s. Considering the trade-off between performance and inference time, we choose 50 inference steps for the final model configuration, as it strikes a good balance between improved performance and maintaining high efficiency.

\subsection{Impact of the Number of Blocks}
We evaluate the impact of the number of Joint-level Interaction Modules in the EJIM on performance. As shown in the \cref{tab:block}, both increasing and decreasing the number of modules lead to a decline in performance. The model with six modules achieves the best balance across all metrics, so we choose this configuration as the final model.

\section{Limitations}\label{limitations}
Although our EJIM method generates realistic HOI animations, it has certain limitations. It faces challenges with multi-object interactions, which can result in incomplete or inaccurate animations. Moreover, it cannot handle non-rigid objects (e.g., hair or smoke) due to the absence of necessary deformation priors in the framework. While incorporating these priors could improve the model’s applicability to complex scenarios, the lack of sufficient data remains a barrier.

\end{document}